\pdfoutput=1

\documentclass[11pt]{article}

\usepackage[final]{acl}

\usepackage{times}
\usepackage{latexsym}
\usepackage{amsmath} 
\usepackage{amsfonts}
\usepackage{graphicx}
\usepackage{adjustbox}
\usepackage{bm}
\usepackage{bbding}
\usepackage{pifont}

\usepackage{amssymb}
\usepackage{wasysym}
\usepackage{graphicx}
\usepackage{subfig}
\usepackage{microtype}
\usepackage{multirow}
\usepackage{amsmath}
\usepackage{amsfonts}
\usepackage{booktabs}
\usepackage{verbatim}
\usepackage{amssymb}
\usepackage{enumitem}
\usepackage[section]{placeins}
\usepackage{float}
\usepackage[T1]{fontenc}

\usepackage[utf8]{inputenc}

\usepackage{microtype}

\usepackage{inconsolata}
\usepackage{hyperref}

\title{Simple but Effective Compound Geometric Operations for Temporal Knowledge Graph Completion}

\author{Rui Ying\textsuperscript{1} \quad 
Mengting Hu\textsuperscript{2}
\thanks{\; Mengting Hu is the corresponding author.} 
\quad 
Jianfeng Wu\textsuperscript{1} \quad 
Yalan Xie\textsuperscript{1} \quad
Xiaoyi Liu\textsuperscript{1} \\
\textbf{Zhunheng Wang\textsuperscript{1} \quad 
Ming Jiang\textsuperscript{1} \quad
Hang Gao\textsuperscript{3} \quad 
Linlin Zhang\textsuperscript{4} \quad 
Renhong Cheng\textsuperscript{1}} \\
\textsuperscript{1} College of Computer Science, Nankai University,
\textsuperscript{2} College of Software, Nankai University \\
\textsuperscript{3} College of Artificial Intelligence, Tianjin University of Science and Technology \\
\textsuperscript{4} China Automotive Technology and Research Center Co., Ltd. \\
\tt ruiying@mail.nankai.edu.cn, \tt mthu@nankai.edu.cn}

\begin{document}
\maketitle
\begin{abstract}


Temporal knowledge graph completion aims to infer the missing facts in temporal knowledge graphs. Current approaches usually embed factual knowledge into continuous vector space and apply geometric operations to learn potential patterns in temporal knowledge graphs. However, these methods only adopt a single operation, which may have limitations in capturing the complex temporal dynamics present in temporal knowledge graphs. Therefore, we propose a simple but effective method, i.e. TCompoundE, which is specially designed with two geometric operations, including time-specific and relation-specific operations. We provide mathematical proofs to demonstrate the ability of TCompoundE to encode various relation patterns. Experimental results show that our proposed model significantly outperforms existing temporal knowledge graph embedding models. Our code  is available at \href{https://github.com/nk-ruiying/TCompoundE}{https://github.com/nk-ruiying/TCompoundE}.

\end{abstract}

\section{Introduction}

A knowledge graph (KG) comprises a collection of structured knowledge presented in triples, offering a simple and effective means of describing factual information \citep{Tiwari2021RecentTI, Liu2023AutomaticKE}. In a KG, a triplet is conventionally represented as $(s, \hat{r}, o)$, where $s$, $o$ and $\hat{r}$ correspond to the head entity, tail entity and the relation linking between the head and tail entities, respectively. However, knowledge is not static in the real world \citep{Li2023TeASTTK}. Temporal knowledge graph (TKG) represents knowledge $(s, \hat{r}, o)$ occurring at timestamp $\tau$, denoted as a quadruple $(s, \hat{r}, o, \tau)$, thereby adding a temporal dimension to knowledge graphs. The introduction of the timestamp $\tau$ enables TKG to delineate the temporal scope of knowledge more accurately and helps us better uncover the potential information within it \citep{Zhang2022AlongTT}. Therefore, TKG is widely applied in downstream tasks such as question answering \citep{Jia2018TEQUILATQ}, information retrieval \citep{Campos2014SurveyOT}, and recommendation systems \citep{Lathia2010TemporalDI} due to its temporal characteristics.



However, TKG usually does not cover all the facts. The incompleteness of TKG hinders the performance of its downstream tasks.
To enhance the overall completeness of TKG, the temporal knowledge graph embedding (TKGE) model utilizes existing knowledge to predict and estimate missing facts. Specifically, the TKGE model employs distinct score functions to acquire effective vector space representations for entities, relations and timestamps, thus utilizing these representations to predict missing facts in TKG. Furthermore, how to enhance the expressive capabilities of the TKGE model is also an important issue that has received widespread attention.

\begin{figure}[t]
    \centering 
    \includegraphics[scale=0.18]{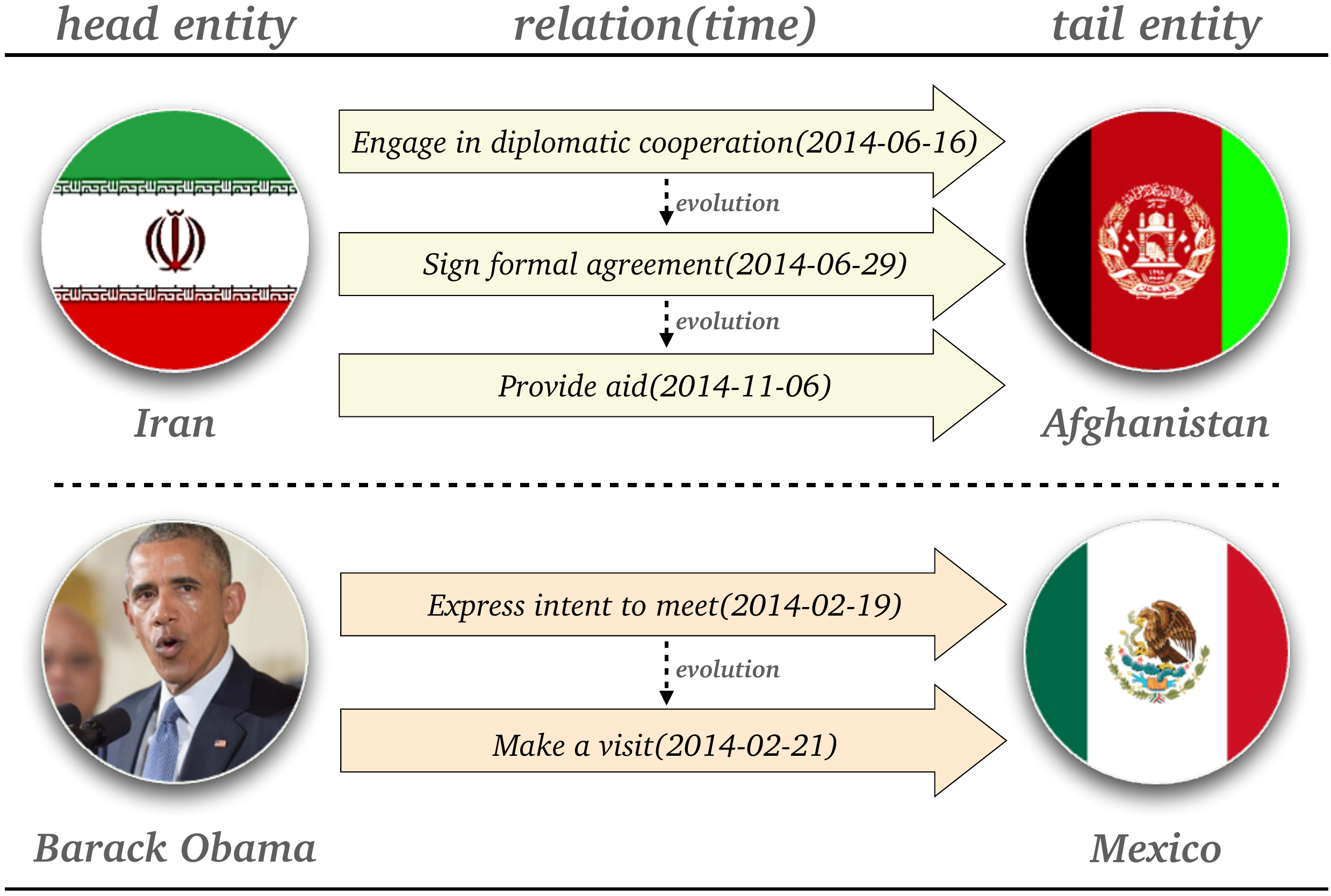}
    \caption{An illustration of temporal evolution patterns. It can be observed that the relationships between head entity and tail entity are dynamically determined by both relation and time.}
    \label{fig:1}
\end{figure}



Geometric operations such as translation and scaling are widely used operations in the field of graphics. These operations help to distinguish between these different classes of entities and to model different relational patterns. Models like TTransE \citep{Leblay2018DerivingVT}, TComplEx \citep{Lacroix2020TensorDF} and TNTComplEx \citep{Lacroix2020TensorDF} all use these operations and achieve good performance. Previous TKGEs often use a single type of operation for embedding. This approach is not conducive to modeling different relation patterns and temporal evolution patterns (shown in Fig. \ref{fig:1})  since each operator may have modeling limitations. \citet{Ge2022CompoundEKG} show that using complex geometry operations in knowledge graphs can effectively model different relational patterns. Inspired by them, we work on compound geometric operations that fit relationships and timestamps in TKGs.

In this paper, we present a model called \textbf{TCompoundE} based on CompoundE \citep{Ge2022CompoundEKG}. Different from CompoundE's work, we discuss the effects of representing relations and timestamps as different geometric operations, which we call relation-specific operations and time-specific operations. More specifically, we use composite operations involving translation and scaling as relation-specific operations and time-specific operations. We integrate time-specific operations within the framework of relation-specific operations, aiming to capture both time-varying and time-invariant features within the low-dimensional representations of relations.  Relation-specific operations, incorporating temporal information, are applied to the head entity embedding. Subsequently, the confidence of the quadruple is expressed by computing the semantic similarity between the head entity embedding and the tail entity embedding.

In summary, the main contributions of our work are as follows:

\begin{itemize}
    \item We present a novel TKG embedding model called TCompoundE, which introduces relation-specific and time-specific compound geometric operations. 
    \item We substantiate the suitability of our model for important relation patterns through mathematical formulations.
    \item 
    Our experimental results on three benchmark datasets demonstrate that our method both meets and surpasses the performance of existing TKGE methods.
\end{itemize}

\section{Related Work}
\subsection{Knowledge Graph Embedding}
Alternatively referred to as static knowledge graph embedding, knowledge graph embedding commonly involves embedding entities and relations into low-dimensional vector spaces. This low-dimensional embedding is employed to enhance the representation of entities and relations through score functions. The model can be categorized into a translation model and a bilinear model based on the distinct score functions employed. \textbf{In translation model}, TransE \citep{Bordes2013TranslatingEF} originally introduces a straightforward and efficient score function $\phi (s,\hat{r},o) = \| \bm{e_s} + \bm{e_{\hat{r}}} - \bm{e_o} {\|}_2$, where $\bm{e_s},\bm{e_{\hat{r}}},\bm{e_o}$ represent the lower dimensional embedding of head entity, relation and tail entity respectively, utilizing relations to denote the translation distance from head entities to tail entities. Subsequent models, including TransH \citep{Wang2014KnowledgeGE}, TransR \citep{Lin2015LearningEA} and TransD \citep{Ji2015KnowledgeGE} adopt distinct projection strategies to implement a relation-specific representation of entities. RotatE \citep{Sun2018RotatEKG} introduces the concept of relations into knowledge graph embedding through a rotation operation in complex space. The PairRE \citep{Chao2020PairREKG} model suggests implementing separate relation-specific scaling operations on the head and tail entities.  HAKE \citep{zhang2020learning} operates by mapping the embeddings into polar coordinate space. CompoudE \citep{Ge2022CompoundEKG} employs a combination of translation, scaling and rotation to form a low-dimensional representation of the relation, addressing the limitations associated with individual operations. \textbf{In bilinear model}, DisMult \citep{Yang2014EmbeddingEA} represents relationships through a symmetric matrix, deriving a scoring function $\phi (s,\hat{r},o) =  <\bm{e_s}, \bm{W_{\hat{r}}}, \bm{e_o} >$  by evaluating the semantic similarity between head and tail entities, where $\bm{W_{\hat{r}}}$ represents the symmetric matrix of relation. ComplEx \citep{Trouillon2016ComplexEF} operates within complex spaces. TuckER \citep{Balazevic2019TuckERTF} employs Tucker decomposition to assess the plausibility of fact triples.

\subsection{Temporal Knowledge Graph Embedding}
Temporal knowledge graph embedding closely resembles static knowledge graph embedding. Therefore, the majority of TKGE are adapted from knowledge graph embeddings to accommodate dynamic changes in facts. For instance, TTransE \citep{Leblay2018DerivingVT}  represents timestamp through a translation operation, following the concept introduced by TransE \citep{Bordes2013TranslatingEF}. Its score function is denoted by $\phi (s,\hat{r},o, \tau) = \| \bm{e_s} + \bm{e_{\hat{r}}} + \bm{e_{\tau}}- \bm{e_o} {\|}_2$, where $\bm{e_{\tau}}$ represents the lower-dimensional embedding of timestamp. TA-DistMult \citep{GarcaDurn2018LearningSE} integrates the timestamp through the application of the DistMult \citep{Yang2014EmbeddingEA} method. TComplEx \citep{Lacroix2020TensorDF} and TNTComplEx \citep{Lacroix2020TensorDF} expand the tensor decomposition methodology introduced by the ComplEx \citep{Trouillon2016ComplexEF} model to encompass four dimensions, facilitating the embedding of TKGs. ChronoR \citep{Sadeghian2021ChronoRRB} employs the rotation operation from RotatE \citep{Sun2018RotatEKG} to comprehend the influence of the combination between timestamp and relation on the entity's rotation operation. RotateQVS \citep{Chen2022RotateQVSRT} employs QuatE \citep{Zhang2019QuaternionKG} quaternion rotation to depict the temporal impact on entities. BoxTE \citep{Messner2021TemporalKG} embeds TKGs using the box embedding model, BoxE \citep{Abboud2020BoxEAB}. \textbf{In addition to this}, DE-Simple \citep{Goel2019DiachronicEF} incorporates diachronic embedding, expanding the static knowledge graph embedding methods into TKGE models capable of handling temporal information. TeAST \citep{Li2023TeASTTK} employs timestamp as a mapping operation for the Archimedean spiral axis, associating relations with the spiral axis to capture the dynamic evolution of relations over time.

\section{Background and Notation}


In subsequent sections of this article, the CompoundE \citep{Ge2022CompoundEKG} methodology is employed to elucidate translation, scaling, and rotation operations. The ensuing discussion provides a comprehensive account of the representation of these operations, followed by an introduction to the CompoundE model.

\subsection{Translation, Rotation, and Scaling}
\begin{figure}
    \centering
    \includegraphics[scale=0.4]{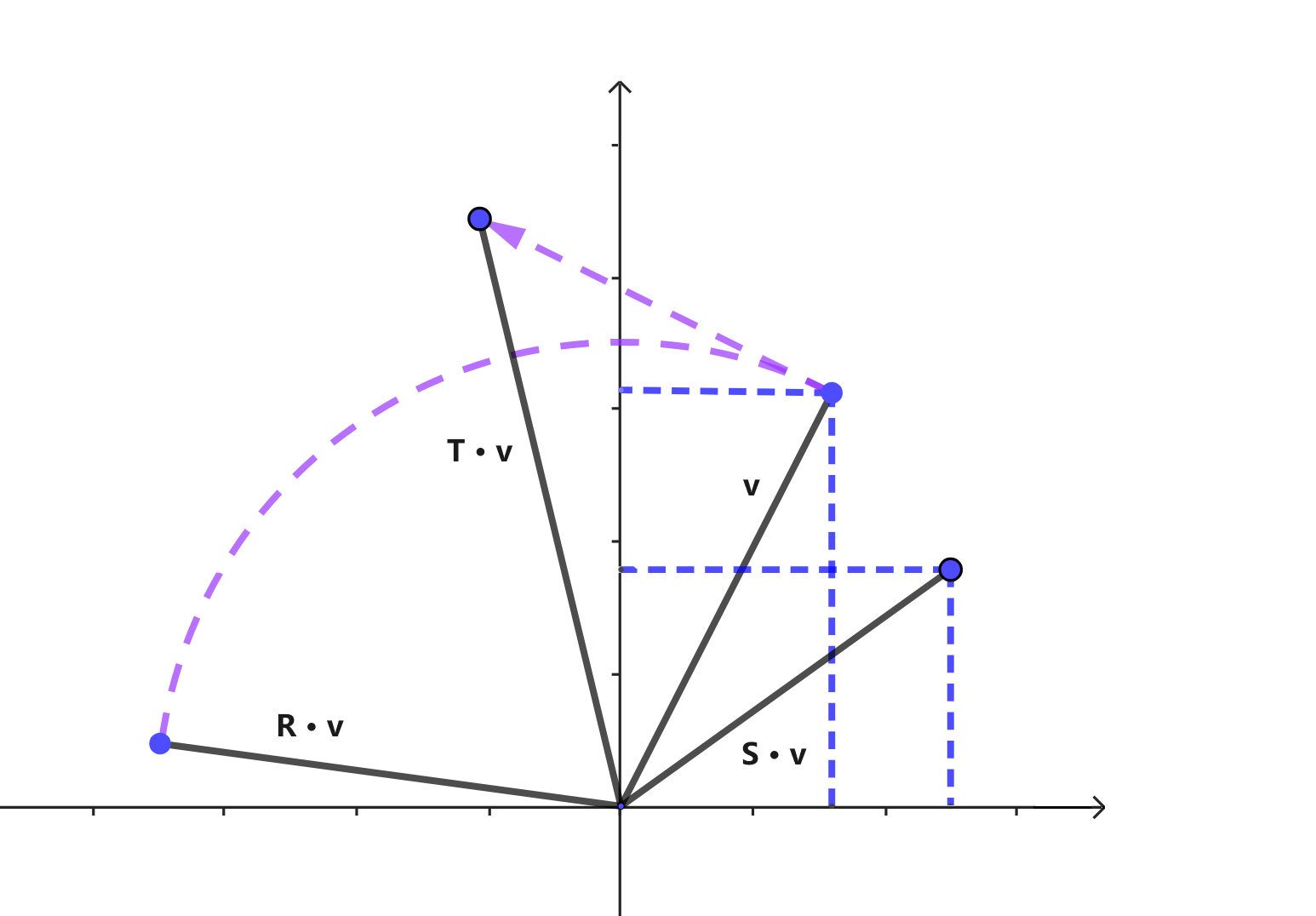}
    \caption{ An illustration of translation, rotation and scaling. Where $\bm{v}$ is the original vector, $\bm{T \cdot v}$ is the vector after the translation operation, $\bm{R \cdot v}$ is the vector after the rotation operation, and $\bm{S \cdot v}$ is the vector after scaling.}
    \label{fig:tsr}
\end{figure}


Translation, rotation and scaling transformations are fundamental operations in graphics, frequently employed in various engineering applications. In the processing of robot motion \citep{RobotMotation}, the cascade application of translation, rotation and scaling operations constitutes a method for precise position determination. 
The translation, rotation and scaling operations are illustrated in Fig. \ref{fig:tsr}. 
Expressing these operations in matrix form proves to be both more efficient and straightforward. 
The translation operation in Fig. \ref{fig:tsr} is represented in 2D as follows:
\begin{equation}\label{...}
    \bm{T} = \left[ \begin{array}{ccc} 1 & 0 & t_x \\ 0 & 1 & t_y \\ 0 & 0 & 1
    \end{array} \right]
\end{equation}
\noindent
while 2D rotation matrix can be written as:
\begin{equation}\label{...}
    \bm{R} = \left[ \begin{array}{ccc} \cos{\theta} & - \sin{\theta} & 0 \\ \sin{\theta} & \cos{\theta} & 0 \\ 0 & 0 & 1
    \end{array} \right]
\end{equation}
\noindent
And 2D scaling matrix can be expressed as:
\begin{equation}\label{...}
    \bm{S} = \left[ \begin{array}{ccc} s_x & 0 & 0 \\ 0 & s_y & 0 \\ 0 & 0 & 1
    \end{array} \right]
\end{equation}


\subsection{CompoundE}
 CompoundE \citep{Ge2022CompoundEKG} encompasses a range of models that integrate various translation, rotation and scaling operations. In this discussion, we focus on a specific combination approximating our model. This involves applying a blend of translation, rotation and scaling to the head entity. The ultimate score is determined by computing the distance between the transformed head and the initial tail entity embeddings. The corresponding score function is formally defined as follows:
\begin{equation}\label{...}
\phi (s,\hat{r},o) =\|      \bm{T_{\hat{r}}} \cdot \bm{R_{\hat{r}}} \cdot \bm{S_{\hat{r}}} \cdot \bm{e_s} - \bm{e_o} \|
\end{equation}
\noindent
where $\bm{S_{\hat{r}}}$, $\bm{R_{\hat{r}}}$, $\bm{T_{\hat{r}}}$ denote the translation, rotation and scaling operations for the head entity embedding. These constituent operators are specific to relations. It is essential to highlight that the scaling, rotation and translation operations employed here are sequential, and altering their order yields distinct outcomes \citep{Ge2022CompoundEKG}.

\subsection{Problem formulation}
For a temporal knowledge graph $\mathcal{G}$, we use $\mathcal{E}$ to denote the set of entities. $\mathcal{R}$ and $\mathcal{T}$ represent the set of relations and the set of timestamps in the temporal knowledge graph respectively. A fact in the temporal knowledge graph is represented by the quadruple $(s, \hat{r}, o, \tau)$, where $s, o \in \mathcal{E} $, $ \hat{r} \in \hat{\mathcal{R}} $ and $ \tau \in \mathcal{T} $. The task of knowledge graph completion is to predict the missing facts through the existing facts in the knowledge graph. We train the data in the training set through the score function $ \phi (s,\hat{r},o, \tau)$ in the model. When predicting a fact, we give either the head entity, relation and timestamp or the tail entity, relation and timestamp in the quadruple. We input these missing quadruples $(s, \hat{r}, ?, \tau)$ or $( ?, \hat{r}, o, \tau)$ and candidate entities into the score function, taking the highest score to form a new fact quadruple for the entity.

\section{Methodology}
\subsection{TCompoundE Model}
In this section, we present our model, TCompoundE, which employs compound geometric operations on both relations and timestamps. For a quadruple $(s, \hat{r}, o, \tau )$ in TKG. We utilize the notations $\bm{e_s}, \bm{e_o}$ to represent the embeddings of the head entity $s$ and tail entity $o$. We utilize $\bm{S_{\hat{r}}}$ and $\bm{T_{\hat{r}}}$ to represent relation-specific scaling, translation  operations. Integrate temporal information into relation-specific operations before applying them to entity embeddings. This merging involves time-specific translation $\bm{T_{\tau}}$ and scaling $\bm{S_{\tau}}$ operations in a relationship-specific process. In our model, we employ translation and scaling operations to represent relation-specific operations and time-specific operations. To facilitate a comprehensive introduction to our model, we categorize it into two distinct sections: \textbf{Time-Specific Operation} and \textbf{Relation-Specific Operation}; The initial section elucidates the utilization of time-specific operations in conjunction with relation-specific operations within a quadruple. In the subsequent section, we elaborate on the impact of relation-specific operations on the embedding of the head entity.


\textbf{Time-Specific Operation}. In our model, we employ time-specific translation $\bm{T_{\tau}}$ and scaling $\bm{S_{\tau}}$ operations to imbue temporal information into the relation. We exclusively apply these operations to the relation-specific scaling operation $\bm{S_{\hat{r}}}$. It is crucial to highlight that we scale the relation-specific operation by first applying translation and then scaling. This sequencing is intentional, as the order of operations can influence the outcome. However, for the relation-specific translation operation, we refrain from integrating time information. This approach aims to capture features of relations that remain constant over time. Subsequently, we obtain relation-specific operations that integrate temporal information. Herein, $\bm{S_{\hat{r} \tau}}$ and $\bm{T_{\hat{r} \tau}}$ denote the relation-specific scaling and translation operations, respectively, after incorporating time information. These operations can be precisely described by the following formula: 
\begin{align}\label{forva}
    \bm{S_{\hat{r} \tau}} &=      \bm{S_{\tau}} \cdot   \bm{T_{\tau}} \cdot \bm{S_{\hat{r}}} \\ 
    \label{forvaT}
    \bm{T_{\hat{r} \tau}} &= \bm{T_{\hat{r}}}
\end{align}
\textbf{Relation-Specific Operation}. We denote the relation-specific translation and relation-specific scaling operations for the head entity as $\bm{T_{\hat{r}}}$ and $\bm{S_{\hat{r}}}$ respectively. To capture temporal information within the TKG, we refrain from applying these operations directly to the head entity embeddings. Instead, we execute relation-specific operations subsequent to the time-specific operations. Specifically, we utilize $\bm{S_{\hat{r} \tau}}$ and $\bm{T_{\hat{r} \tau}}$ to conduct relation-specific operations incorporating time information on the head entity embedding. This operation is formally represented as:
\begin{equation}\label{headevo}
\bm{e_{s}^{\hat{r} \tau}} = \bm{S_{\hat{r} \tau}} \cdot   \bm{T_{\hat{r} \tau}}  \cdot \bm{e_s}
\end{equation}

We obtain the head entity representation $\bm{e_{s}^{\hat{r} \tau}}$ incorporating fused time and relation information using Formula \ref{headevo}. Unlike CompoundE \citep{Ge2022CompoundEKG}, our chosen score function is not a distance metric; instead, it is determined by the semantic similarity between $\bm{e_{s}^{\hat{r} \tau}}$ and the tail entity $\bm{e_o}$. This decision is grounded in our belief that semantic similarity offers more advantages than distance metrics in the context of TKG \emph{(proof in Appendix \ref{apx:difscore})}. The score function for TCompoundE is expressed as:
\begin{equation}\label{fuc:sim}
\phi (s,\hat{r},o, \tau) = <\bm{e_{s}^{\hat{r} \tau}}, \bm{e_o}>
\end{equation}

\subsection{Loss Function}
Building upon TNTComplEx \citep{Lacroix2020TensorDF} and TeAST \citep{Li2023TeASTTK}, we adopt reciprocal learning for training our model, with the loss function defined as follows:
\begin{large}
\begin{equation}\label{...}
\begin{split}
\mathcal{L}_u = -  \log( \frac{\exp (\phi (s,\hat{r},o, \tau))}{\sum_{o' \in \varepsilon } \exp (\phi (s,\hat{r},o', \tau))}  ) \\
 - \log( \frac{\exp (\phi (o,\hat{r}^{-1},s, \tau))}{\sum_{s' \in \varepsilon } \exp (\phi (o,\hat{r}^{-1},s', \tau))}  )\\
 + \lambda_u\sum\limits_{i=1}^{k} (\| \bm{e_s} \|_3^3 + \| \bm{T_{\hat{r} \tau}} + \bm{S_{\hat{r} \tau}} \|_3^3 + \| \bm{e_o} \|_3^3 )
\end{split}
\end{equation}
\end{large}
\noindent
where $\lambda_u$ is the weight of the N3 regularization, and $r^{-1}$ denotes the inverse relation. We use the smoothing temporal regularizer in TNTComplEx \citep{Lacroix2020TensorDF} to make neighboring timestamps having close representations. It is defined as:
\begin{equation}\label{...}
\mathcal{L}_{\tau} = \frac{1}{N_{\tau - 1}} \sum\limits_{i=1}^{N_{\tau}-1} \| \bm{e_{\tau(i+1)} - e_{\tau(i)}} \|_3^3
\end{equation}

The total loss function of TCompoundE is defined as:
\begin{equation}\label{...}
\mathcal{L} = \mathcal{L}_u + \lambda_{\tau} \mathcal{L}_{\tau}
\end{equation}
where $\lambda_{\tau}$ represents temporal regularization.

\subsection{Modeling Various Relation Patterns}
TCompoundE can model crucial relation patterns, encompassing symmetric, asymmetric, inverse and temporal evolution patterns (detail in Appendix \ref{appendixA}). We enumerate all the propositions in this section, with corresponding proofs provided in the Appendix.
\newtheorem{proposition}{Proposition}
\begin{proposition}
TCompoundE can model the symmetric relation pattern. (proof in Appendix \ref{proof:1})
\end{proposition}
\begin{proposition}
TCompoundE can model the asymmetric relation pattern. (proof in Appendix \ref{proof:2})
\end{proposition}
\begin{proposition}
TCompoundE can model the inverse relation pattern. (proof in Appendix \ref{proof:3})
\end{proposition}
\begin{proposition}
TCompoundE can model the temporal evolution relation pattern. (proof in Appendix \ref{proof:4})
\end{proposition}

\section{Experiments}
\label{sec:exp}
\subsection{Datasets}
\begin{table}[ht]
\small
    \centering
    \setlength{\tabcolsep}{1.2mm}{
    \begin{tabular}{l|c|c|c}
    \toprule
     & {$\mathtt{ICEWS14}$} & {$\mathtt{ICEWS05-15}$} & {$\mathtt{GDELT}$} \\
    \midrule
    \#${\mathcal{E}}$ & 7,128 & 10,488 & 500 \\
    \#${\hat{\mathcal{R}}}$ & 230 &  251 & 20 \\
    \#${\mathcal{T}}$ & 365 & 4017 & 366 \\
    \midrule
    \#Train & 72,826 & 386,962 & 2,735,685 \\
    \#Dev & 8,963 & 46,092 & 31,961 \\
    \#Test & 8,941 & 46,275& 31,961 \\
\bottomrule
    \end{tabular}}
    \caption{Data statistics. \#${\mathcal{E}}$, \#${\hat{\mathcal{R}}}$ and \#${\mathcal{T}}$ denote the number of entities, the number of relations and the number of timestamps respectively.}
    \label{table:data}
\end{table}
We assess the performance of TCompoundE on three benchmark datasets for TKGE. The datasets include ICEWS14 and ICEWS05-15 \citep{GarcaDurn2018LearningSE}, both derived from the Integrated Crisis Early Warning System (ICEWS) \citep{icews}. The ICEWS database system records significant political events, with ICEWS14 capturing events from the year 2014, and ICEWS05-15 encompassing major political events spanning from 2005 to 2015. Additionally, we evaluate TCompoundE on GDELT \citep{gdelt}, a subset of the extensive Global Database of Events, Language, and Tone (GDELT) Temporal Knowledge Graph dataset. GDELT incorporates information from diverse sources, comprising facts with daily timestamps between April 1, 2015, and March 31, 2016. Notably, GDELT is limited to the 500 most common entities and 20 most frequent relations. More information about the datasets can be found in Table \ref{table:data}.
\subsection{Baselines}
We conduct a comprehensive comparison of our model with state-of-the-art Temporal Knowledge Graph Embedding (TKGE) models, which include TTransE \citep{Leblay2018DerivingVT}, DE-SimplE \citep{Goel2019DiachronicEF}, TA-DisMult \citep{GarcaDurn2018LearningSE}, ChronoR \citep{Sadeghian2021ChronoRRB}, TComplEx \citep{Lacroix2020TensorDF}, TNTComplEx \citep{Lacroix2020TensorDF}, BoxTE \citep{Messner2021TemporalKG}, RotateQVS \citep{Chen2022RotateQVSRT}, and TeAST \citep{Li2023TeASTTK}.

Among these existing TKGE methods, TeAST achieves state-of-the-art results on the ICEWS14, ICEWS05-15, and GDELT datasets. Consequently, we consider TeAST \citep{Li2023TeASTTK} as the primary baseline for our comparative analysis. In addition, we also used a variant of the TCompoundE model for ablation experiments. A detailed description of the variant model can be found in Appendix \ref{apd:var}. 

\begin{table*}[ht]
\small
    \centering
    \setlength{\tabcolsep}{2mm}{
    \begin{tabular}{l|cccc|cccc|cccc}
    \toprule
    \multirow{1}{*}{Methods} & \multicolumn{4}{c|}{$\mathtt{ICEWS14}$} & \multicolumn{4}{c|}{$\mathtt{ICEWS05-15}$} & \multicolumn{4}{c}{$\mathtt{GDELT}$} \\
    & $\mathtt{MRR}$ & $\mathtt{H@1}$ & $\mathtt{H@3}$ & $\mathtt{H@10}$ & $\mathtt{MRR}$ & $\mathtt{H@1}$ & $\mathtt{H@3}$ & $\mathtt{H@10}$ & $\mathtt{MRR}$ & $\mathtt{H@1}$ & $\mathtt{H@3}$ & $\mathtt{H@10}$ \\
    \midrule
    {\small TTransE} & {\small 0.255} & {\small 0.074} & {\small -} & { \small 0.601} & {\small 0.271} & {\small 0.084} & {\small -} & { \small 0.616} & {\small 0.115} & {\small 0.0} & {\small 0.160} & { \small 0.318}  \\
{\small DE-SimplE} & {\small 0.526} & {\small 0.418} & {\small 0.592} & { \small 0.725} & {\small 0.513} & {\small 0.392} & {\small 0.578} & { \small 0.748} & {\small 0.230} & {\small 0.141} & {\small 0.248} & { \small 0.403}  \\
{\small TA-DisMult} & {\small 0.477} & {\small 0.363} & {\small -} & { \small 0.686} & {\small 0.474} & {\small 0.346} & {\small -} & { \small 0.728} & {\small 0.206} & {\small 0.124} & {\small 0.219} & { \small 0.365}  \\
{\small ChronoR} & {\small 0.625} & {\small 0.547} & {\small 0.669} & { \small 0.773} & {\small 0.675} & {\small 0.596} & {\small 0.723} & { \small 0.820} & {\small -} & {\small -} & {\small -} & { \small -}  \\
{\small TComplEx} & {\small 0.610} & {\small 0.530} & {\small 0.660} & { \small 0.770} & {\small 0.660} & {\small 0.590} & {\small 0.710} & { \small 0.800} & {\small 0.340} & {\small 0.294} & {\small 0.361} & { \small 0.498}  \\
{\small TNTComplEx} & {\small 0.620} & {\small 0.520} & {\small 0.660} & { \small 0.760} & {\small 0.670} & {\small 0.590} & {\small 0.710} & { \small 0.810} & {\small 0.349} & {\small 0.258} & {\small 0.373} & { \small 0.502}  \\
{\small BoxTE} & {\small 0.613} & {\small 0.528} & {\small 0.664} & { \small 0.763} & {\small 0.667} & {\small 0.582} & {\small 0.719} & { \small 0.820} & {\small 0.352} & {\small 0.269} & {\small 0.377} & { \small 0.511}  \\
{\small RotateQVS} & {\small 0.591} & {\small 0.507} & {\small 0.642} & { \small 0.754} & {\small 0.633} & {\small 0.529} & {\small 0.709} & { \small 0.813} & {\small 0.270} & {\small 0.175} & {\small 0.293} & { \small 0.458}  \\
{\small TeAST} & {\small 0.637} & {\small 0.560}  & { \small 0.682} & {\small 0.782} & {\small 0.683} & {\small 0.604} & { \small 0.732} & {\small 0.829} & {\small 0.371}& {\small 0.283} & {\small 0.401} & { \small 0.544}  \\

\midrule
{\small TCompoundE} & {\textbf{ \small 0.644}} & {\textbf{\small 0.561}} & {\textbf{\small 0.694}} & { \textbf{\small  0.795}} & {\textbf{\small 0.692}} & {\textbf{\small 0.612}} & {\textbf{\small 0.743}} & { \textbf{\small 0.837}} & {\textbf{\small 0.433}} & {\textbf{\small 0.347}} & {\textbf{\small 0.469}} & {\textbf{ \small 0.595}}  \\
\bottomrule
    \end{tabular}
    }
    \caption{Knowledge graph completion results on ICEWS14, ICEWS05-15 and GDELT. ALL results are taken from the original papers.}
    \label{table:main}
\end{table*}
\subsection{Evaluation Protocol}
This paper evaluates our temporal knowledge graph embedding (TKGE) model using the aforementioned benchmarks. Following established baselines, we assess the ranking quality of each test triplet by computing all possible substitutions for both the head and tail entities. Specifically, for a missing quadruple $(s, \hat{r}, ?, \tau)$ or $(?, \hat{r}, o, \tau)$, we calculate the scores for all entities and rank them accordingly. We then filter out other correct quadruples to make predictions about the current missing entity. Performance is evaluated using standard metrics, including mean reciprocal rank (MRR) and Hits@n. Hits@n measures the percentage of correct entities within the top n predictions. Higher values of MRR and Hits@n indicate superior performance. The Hits ratio is calculated with cut-off values n = 1, 3, 10, denoted as H@n for convenience in this paper.

\subsection{Experimental Setup}
We implement our proposed model TCompoundE via pytorch based on the TeAST training framework. All experiments are trained on a single NVIDIA RTX A6000 with 48GB memory. We use the Adagrad optimizer and conduct a lot of experiments to find the optimal parameter configuration on each dataset. The learning rate is set to 0.01 and the embedding dimension $d$ is set to 6000 and the batch size is set to 4000 in the ICEWS14 dataset and the max epoch is set to 400. In the ICEWS05-15 dataset, we set the learning rate to 0.08, $d$ to 8000, the batch size to 6000 and the max epoch to 100. In GDELT, the learning rate, $d$, the batch size and the max epoch are set to 0.35, 6000, 2000 and 50 respectively. The optimal hyperparameters for TCompound are as follows:
\begin{itemize}
    \item \textbf{ICEWS14}: $ \lambda_u = 0.0025, \quad \lambda_{\tau} = 0.01$
    \item \textbf{ICEWS05-15}:$ \lambda_u = 0.002, \quad \lambda_{\tau} = 0.1$
    \item \textbf{GDELT}:$ \lambda_u = 0.001, \quad \lambda_{\tau} = 0.001$
\end{itemize}
We report the average results on the test set for five runs.

\section{Results and Analysis}

\begin{table*}[t]
\small
    \centering
    \begin{tabular}{c|c|ccc|cccc|cccc}
    \toprule
    \multirow{1}{*}{Model} & \multirow{1}{*}{Mapping on} & \multicolumn{3}{c}{$\mathtt{Relation}$} & \multicolumn{4}{c}{$\mathtt{ICEWS14}$}& \multicolumn{4}{c}{$\mathtt{ICEWS05-15}$}  \\
    & & $\bm{T_{\hat{r}}}$ & $\bm{S_{\hat{r}}}$ & $\bm{R_{\hat{r}}}$ & $\mathtt{MRR}$ & $\mathtt{H@1}$ & $\mathtt{H@3}$ & $\mathtt{H@10}$ & $\mathtt{MRR}$ & $\mathtt{H@1}$ & $\mathtt{H@3}$ & $\mathtt{H@10}$ \\
    \midrule
    TCompoundE &$ \bm{S_{\hat{r}}}$  & \ding{52} & \ding{52} & & \textbf{0.644} & \textbf{0.561} & \textbf{0.694} & \textbf{0.795} &                \textbf{0.692} & \textbf{0.612} & \textbf{0.743} & \textbf{0.837}  \\
    \midrule
    V1
    & \multirow{3}{*}{$\bm{T_{\hat{r}}}$}
    & \ding{52} & & & 0.304 & 0.128 & 0.412 & 0.632 & 
                                  0.374 & 0.194 & 0.490 & 0.702 \\
    V2 & & \ding{52} & \ding{52} & & 0.576 & 0.476 & 0.639 & 0.755                                & 0.623 & 0.521 & 0.690 & 0.809\\
    V3 & & \ding{52} & \ding{52} & \ding{52} & 0.571 & 0.468 & 0.636 & 0.757          & 0.620 & 0.517 & 0.686 & 0.807 \\
    \midrule
    V4 & \multirow{2}{*}{ $\bm{S_{\hat{r}}}$}
    & & \ding{52} & & \underline{0.602} & \underline{0.518} & 0.650 & \underline{0.762}
                               & \underline{0.676} & \underline{0.595} & \underline{0.727} & \underline{0.824} \\
    V5 & & \ding{52} & \ding{52} & \ding{52} & 0.576 & 0.485 & 0.630 & 0.746          & 0.652 & 0.571 & 0.702 & 0.801 \\
    \midrule
    V6 & \multirow{2}{*}{ $\bm{R_{\hat{r}}}$}
    & & & \ding{52}  & 0.601 & 0.515 & \underline{0.651} & 0.761       
                                & 0.634 & 0.555 & 0.680 & 0.781 \\
    V7& & \ding{52} & \ding{52} & \ding{52} & 0.582 & 0.491 & 0.638 & 0.752         & 0.638 & 0.550 & 0.690 & 0.804 \\
\bottomrule
    \end{tabular}
    \caption{Results of different relation-specific operations on ICEWS14 and ICEWS05-15.}
    \label{table:v1}
\end{table*}

\begin{table*}[]
\small
    \centering
    \begin{tabular}{c|c|ccc|cccc|cccc}
    \toprule
    \multirow{1}{*}{Model} &\multirow{1}{*}{Mapping on} & \multicolumn{3}{c}{$\mathtt{Time}$} & \multicolumn{4}{c}{$\mathtt{ICEWS14}$}& \multicolumn{4}{c}{$\mathtt{ICEWS05-15}$}  \\
    & & $\bm{T_{\tau}}$ & $\bm{S_{\tau}}$ & $\bm{R_{\tau}}$ & $\mathtt{MRR}$ & $\mathtt{H@1}$ & $\mathtt{H@3}$ & $\mathtt{H@10}$ & $\mathtt{MRR}$ & $\mathtt{H@1}$ & $\mathtt{H@3}$ & $\mathtt{H@10}$ \\
    \midrule
    
    TCompoundE &$ \bm{S_{\hat{r}}} $ & \ding{52} & \ding{52} & & \textbf{0.644} & \textbf{0.561} & \textbf{0.694} & \textbf{0.795} &                \underline{0.692} & \textbf{0.612} & \underline{0.743} & \underline{0.837}  \\
    \midrule
    V8&\multirow{4}{*}{$\bm{T_{\hat{r}}}$}
    & \ding{52} & & & 0.578 & 0.478 & 0.640 & 0.756 
                               & 0.621 & 0.519 & 0.685 & 0.807  \\
    V9 & & & \ding{52} & & 0.577 & 0.478 & 0.639 & 0.754        
                               & 0.620 & 0.517 & 0.685 & 0.807 \\
    V10& & & & \ding{52} & 0.559 & 0.460 & 0.619 & 0.737        
                               & 0.584 & 0.482 & 0.646 & 0.771 \\
    V11& & \ding{52} & \ding{52} & \ding{52} & 0.564 & 0.462 & 0.628 & 0.747                                  & 0.606 & 0.505 & 0.670 & 0.794 \\
    \midrule
    V12&\multirow{4}{*}{$\bm{S_{\hat{r}}}$}
    & \ding{52} & & & 0.634 & 0.548 & \underline{0.687} & \underline{0.791} 
          & \textbf{0.693} & \underline{0.607} & \textbf{0.750} & \textbf{0.846} \\
    V13 & & & \ding{52} & & \underline{0.638} & \underline{0.558} & 0.685 & 0.783         
                               & 0.682 & 0.606 & 0.735 & 0.832 \\
    V14 & & & & \ding{52} & 0.587 & 0.499 & 0.642 & 0.743         
                               & 0.639 & 0.557 & 0.688 & 0.786 \\
    V15 & & \ding{52} & \ding{52} & \ding{52} & 0.582 & 0.505 & 0.627 & 0.734                                  & 0.673 & 0.595 & 0.723 & 0.815  \\
\bottomrule
    \end{tabular}
    \caption{Results of different time-specific operations on ICEWS14 and ICEWS05-15.}
    \label{table:vtime}
\end{table*}
\subsection{Main Results}
Table \ref{table:main} presents the results of knowledge graph completion for ICEWS14, ICEWS05-15 and GDELT datasets. 
The best-performing results are highlighted in bold font.
Our observations indicate that TCompoundE outperforms all baseline models across all metrics for t datasets. 
As TCompoundE exclusively applies the time-specific operation to the relation-specific scaling operation, this enables relations occurring simultaneously to utilize the same time-specific operation, facilitating the evolution of all relations over time. 
For the relation-specific translation operation, we maintain the timestamps unchanged to preserve the features of relations that remain constant over time. 
This demonstrates the significance of applying the time-specific operation exclusively to the relation-specific scaling operation. 
Additionally, Table \ref{table:main} reveals that TCompoundE has achieved significant improvements over TKGE models utilizing a single operation for both relation-specific and time-specific operations, such as translation (TTransE) and scaling (TComplEx). 
This confirms that employing compound geometric operations for both relation-specific and time-specific operations is an effective strategy for designing TKG embeddings. 
Furthermore, BoxTE \citep{Messner2021TemporalKG} highlights that GDELT necessitates a considerable level of temporal inductive capacity to achieve effective encoding. 
This is because GDELT displays a notable degree of temporal variability, wherein certain facts endure across multiple consecutive timestamps, while others are momentary and sparse. The performance of our model demonstrates it can capture the dynamic evolution of relations. 
The results across all three datasets indicate the effectiveness of our combined approach in addressing the temporal knowledge graph completion problem.

\subsection{Different Combinations} 
\label{sec:diff}
We investigate the impact of various combinations on ICEWS14 and ICEWS05-15. And we also compare the performance of TCompoundE with its variants. More comprehensive experimental results can be found in Appendix \ref{Appedix:E}. We classify these variants into two groups: the first group involves maintaining the time-specific operation unchanged while modifying the relation-specific operation, and the second group involves maintaining the relation-specific operation unchanged while modifying the time-specific operation. The results of the ICEWS14 and ICEWS05-15 variants from the aforementioned two groups are presented in Table \ref{table:v1} and Table \ref{table:vtime}. The best results are highlighted in bold font, while the second best are underlined. 

In Table \ref{table:v1}, we can observe that TCompoundE outperforms its variants across all metrics on ICEWS14 and ICEWS05-15. As TCompoundE utilizes translation and scaling operations as relation-specific operations, this enables TCompoundE to model critical relation patterns. We can observe from Table \ref{table:v1} that TCompoundE has significantly outperformed its variants when employing a single operation for relation-specific operations, including translation (V1), scaling (V4) and rotation (V6). It confirms that utilizing a single operation as relation-specific has drawbacks in TKGs. We also observe that TCompoundE outperforms its variants across all geometric operations, including V3, V4 and V7. These variant models employ translation, scaling, and rotation as relation-specific operations. The distinction among these models lies in the usage of time-specific operations in different relation-specific operations, such as translation (V3), scaling (V5) and rotation (V7). This underscores that incorporating too many geometric operations as relation-specific operations is not optimal in TKGs. Conversely, V2 utilizes the same relation-specific operation as TCompoundE, while applying the time-specific operation to the relation-specific translation operation. This indicates that more features that remain constant over time can be learned from relation-specific translation operations.

\begin{figure*}[!t]
\centering
\subfloat[The initial state of $e_s$ and $e_o$]{
		\includegraphics[scale=0.5]{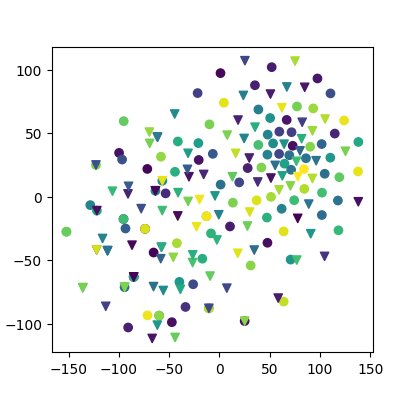}}
\subfloat[$e_s$ apply $T_{\hat{r} \tau}$]{
		\includegraphics[scale=0.5]{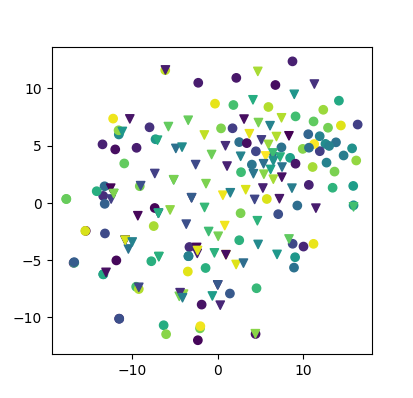}}
\subfloat[$e_s$ apply $T_{\hat{r} \tau}$ and $S_{\hat{r} \tau}$]{
		\includegraphics[scale=0.5]{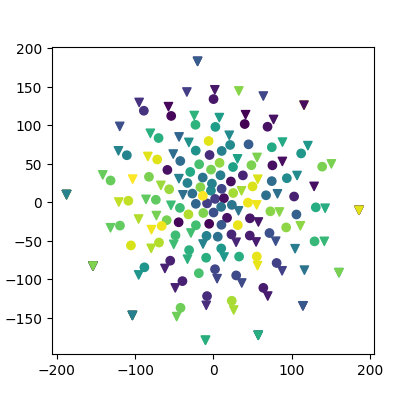}}
\caption{Visualisations of the learned entity embeddings on ICEWS14. The circle in the figure represents the head entity and the triangle represents the tail entity; Circles and triangles of the same color indicate that the head and tail entities come from the same quadruple.}
\label{fig:emb}
\end{figure*}

Table \ref{table:vtime} displays the results of the second group of TCompoundE variants. In Table \ref{table:vtime}, V8, V9, and V10 employ single translation, scaling and rotation as time-specific operations. V12, V13 and V14 are identical to V8, V9 and V10, respectively. From the results, we observe that TCompoundE remains highly competitive compared to its variant models when employing a single operation as a time-specific operation. V11 and V15 use translation, scaling and rotation as time-specific operations, applying them to relation-specific translation and relation-specific scaling operations, respectively. It is evident from Table \ref{table:vtime} that TCompoundE demonstrates significant improvements compared to its variants when employing all geometric operations as time-specific operations. This confirms that incorporating too many operations is often suboptimal.

\subsection{Effects of Operation} \label{sec:sne}
\begin{figure}[t]
    \centering
    \includegraphics[scale=0.35]{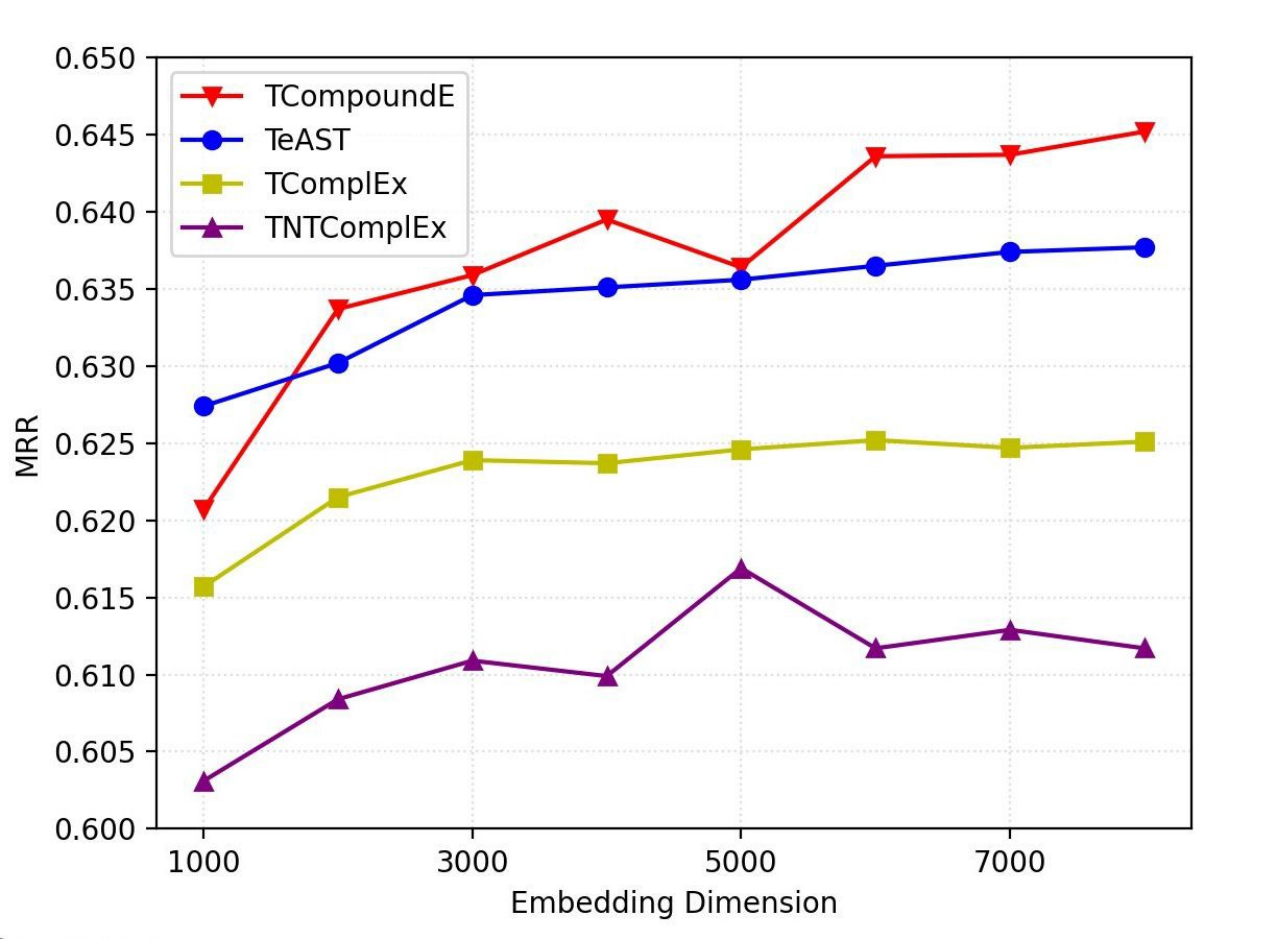}
    \caption{MRR scores on ICEWS14 dataset.}
    \label{fig:comp}
\end{figure}


We randomly select 100 quadruples from ICEWS14 and employ t-SNE \citep{van2008visualizing} to visualize the head and tail entity embeddings of TCompoundE. The visualization results are depicted in Fig. \ref{fig:emb}. We note that the initial state of the head entity and tail entity embeddings appears chaotic and irregular. However, when the head entity embeddings undergo $T_{\hat{r} \tau}$ and $S_{\hat{r} \tau}$ transformations, the distribution of the head and tail entity embeddings becomes more regular. The reason for this phenomenon, we think head entity is mapped to a space close to tail entity by relation-specific operation merging the temporal information. This is due to the design of the scoring function and illustrates the effectiveness of the composite operation. Specifically, the head and tail entities in the same quadruple exhibit a high degree of proximity. This confirms the effectiveness of employing both relation-specific and time-specific operations. In addition, we have a case study of the quadruple we use. Details can be found in the appendix \ref{apx:case}.


\subsection{Effects of Embedding Dimension}
We investigate the effect of embedding dimension on TCompoundE. In Fig. \ref{fig:comp}, we compare the MRR scores of TCompoundE and previous state-of-the-art (SOTA) embedding models on the ICEWS14 dataset across different dimension settings $d \in {1000, 2000, 3000, 4000, 5000, 6000, 7000, 8000}$. According to the experimental results, TCompoundE is suboptimal in low dimensions, but it outperforms benchmarking methods when $d \geq 2000$.

\section{Conclusion}
\label{sec:bibtex}
This paper introduces a novel method, TCompoundE, designed to address the challenge of knowledge graph completion in TKGs. TCompoundE applies a compound of translation and scaling as relation-specific and time-specific operations. Our experimental results demonstrate that TCompoundE effectively manages both relation and time information in TKGs. Furthermore, we provide mathematical evidence supporting TCompoundE's capability to handle various relational patterns. Additionally, we explore the effectiveness of different combinations of relation-specific and time-specific operations.

\section*{Limitations}
Similar to many temporal knowledge graph embedding models, our proposed method, TCompoundE, faces limitations during the training phase as it cannot learn about invisible entities and time. Consequently, TCompoundE cannot directly perform temporal knowledge graph extrapolation tasks. Additionally, our model excels at high embedding dimensions, leading to its larger size compared to others.

\section*{Acknowledgements}
We sincerely thank all the anonymous reviewers for providing valuable feedback. This work is supported by the youth program of National Science Fund of Tianjin, China (Grant No. 22JCQNJC01340).


\bibliography{anthology,custom}

\section*{Appendix}
\label{sec:appendix}
\appendix
\section{Definition of Relation Patterns}\label{appendixA}
\newtheorem{definition}{Definition}
\begin{definition}
A relation $\hat{r}$ is symmetric, if $\forall s,o,\tau, (s,\hat{r},o,\tau) \wedge (o,\hat{r},s,\tau) \in \mathcal{G}$
\end{definition}
\begin{definition}
A relation $\hat{r}$ is asymmetric, if $\forall s,o,\tau, (s,\hat{r},o,\tau) \in \mathcal{G} \wedge (o,\hat{r},s,\tau) \notin \mathcal{G}$
\end{definition}
\begin{definition}
Relation $\hat{r}_{1}$ is the inverse of $\hat{r}_{2}$, if $\forall s,o,\tau, (s,\hat{r}_{1},o,\tau) \wedge (o,\hat{r}_{2},s,\tau) \in \mathcal{G}$
\end{definition}
\begin{definition}
Relation $\hat{r}_{1}$ and $\hat{r}_{2}$ are evolving over time from timestamp $\tau_{1}$ to timestamp $\tau_{2}$, if $\forall s,o, (s,\hat{r}_{1},o,\tau_{1}) \wedge (s,\hat{r}_{2},o,\tau_{2}) \in \mathcal{G}$.
\end{definition}

The above describes various relational patterns from a mathematical point of view. We will illustrate various relational patterns next. For symmetric patterns, \emph{(Canada, Consult, France)} and \emph{(France, Consult, Canada)} show that \emph{Consult} is symmetric. For asymmetric patterns, \emph{is father of} is an asymmetric relation, because \emph{(personA, is father of, personB)} and \emph{(personB, is father of, personA)} can't both be true. \emph{is father of} and \emph{is son of} are inverse relations. Temporal evolution patterns detail can be found at Figure \ref{fig:1}.

\section{Proof of Propositions 1} \label{proof:1}
Let M denote the compound operation for head entity. Following ComplEx, we employ the standard dot product $<a,b>=a \circ b = \sum_k a_k b_k$. For $<a,b> = <c, d>$, there are special cases that the previous equation holds true when $a_ib_i == c_id_i$, where $i \in [0 , k]$. For symmetric pattern, we can get $\phi (s,\hat{r},o, \tau) = \phi (s,\hat{r},o,\tau)$ based on definition of symmetric pattern. According to score function of TCompoundE, we get:
\begin{equation}\label{...}
\begin{split}
<M e_s, e_o> = <M e_o, e_s> \Rightarrow \\
Me_s \circ e_o = M e_o \circ e_s 
\end{split}
\end{equation}
If the matrix M is invertible,then we can get:
\begin{equation}\label{...}
\begin{split}
Me_s \circ e_o = M e_o \circ e_s \Rightarrow \\
e_s \circ e_o = M^{-1}M e_o \circ e_s \Rightarrow\\
e_s \circ e_o = e_o \circ e_s
\end{split}
\end{equation}
Therefore, TCompoundE can model symmetric pattern when matrix M is invertible.

\section{Proof of Propositions 2} \label{proof:2}
By definition of asymmetric pattern, we can get $\phi (s,\hat{r},o, \tau) \neq \phi (s,\hat{r},o,\tau)$. By similar proof for Propositions 1, TCompoundE can model asymmetric pattern when matrix is not invertible.

\section{Proof of Propositions 3} \label{proof:3}
Based on definition of inverse pattern, we have $\phi (s,\hat{r}_1,o, \tau) = \phi (o,\hat{r}_2,s,\tau)$. Hence, we get
\begin{equation}\label{...}
\begin{split}
M_1e_s \circ e_o = M_2 e_o \circ e_s
\end{split}
\end{equation}
If the matrix $M_1$ or $M_2$ is invertible, the we can get:
\begin{equation}\label{...}
\begin{split}
e_s \circ e_o = M_1^{-1}M_2 e_o \circ e_s \Rightarrow \\
M_1^{-1}M_2 = I \quad or \quad M_2^{-1}M_1 = I
\end{split}
\end{equation}
Therefore, TCompoundE can model symmetric pattern when $M_1$ and $M_2$ are inverse matrices.

\section{Proof of Propositions 4} \label{proof:4}
For temporal evolution pattern, we can get $\phi (s,\hat{r}_1,o, \tau_1) = \phi (s,\hat{r}_2,o,\tau_2)$ based on definiton of temporal evolution pattern. Then through the score function, we can get:
\begin{equation}\label{...}
\begin{split}
M_{\hat{r}_1 \tau_1}e_s \circ e_o = M_{\hat{r}_2 \tau_2} e_o \circ e_s
\end{split}
\end{equation}
where $M_{\hat{r}_1 \tau_1}=S_{\tau_1}T_{\tau_1}S_{\hat{r}_1}T_{\hat{r}_1}$ and $M_{\hat{r}_2 \tau_2}=S_{\tau_2}T_{\tau_2}S_{\hat{r}_2}T_{\hat{r}_2}$. Then we can get:
\begin{equation}\label{...}
\begin{split}
e_s \circ e_o = M^{-1}_{\hat{r}_1 \tau_1} M_{\hat{r}_2 \tau_2} \circ e_s \Rightarrow \\
 M^{-1}_{\hat{r}_1 \tau_1} M_{\hat{r}_2 \tau_2} = I \quad or \quad  M^{-1}_{\hat{r}_2 \tau_2} M_{\hat{r}_1 \tau_1} = I
\end{split}
\end{equation}
We can observe from the above formula: TCompoundE can model temporal evolution pattern when $M_{\hat{r}_1 \tau_1}$ and $ M_{\hat{r}_2 \tau_2}$ are inverse matrices.


\section{Introduction of Variant Models}
\label{apd:var}
We introduce $\bm{R_{\hat{r}}}$ and $\bm{R_{\tau}}$ in the variant model to represent the relation-specific and time-specific rotation operation. To better explain the differences between variant models, we introduce a more general formula instead of Formula \ref{headevo}:
\begin{equation}
    \bm{e^{\hat{r} \tau}_s} = \bm{R_{\hat{r} \tau}} \cdot \bm{S_{\hat{r} \tau}} \cdot \bm{T_{\hat{r} \tau}} \cdot \bm{e_s}
\end{equation}
where $\bm{R_{\hat{r} \tau}}$ represents relation-specific rotation operation that incorporates temporal information. Next we will look at the formula differences between TCompoundE and its variant models in obtaining the $\bm{R_{\hat{r} \tau}}$, $\bm{S_{\hat{r} \tau}}$ and $\bm{T_{\hat{r} \tau}}$.

\begin{table}[ht]
\small
    \centering
    \renewcommand{\arraystretch}{1.5}
    \setlength{\tabcolsep}{1.2mm}{
    \begin{tabular}{c|c|c|c}
    \toprule
     & {$\bm{R_{\hat{r} \tau}}$ } &  {$\bm{S_{\hat{r} \tau}}$ } & {$\bm{T_{\hat{r} \tau}}$}  \\
    \midrule
    TCompoundE 
    & $\bm{I}$ 
    & $ \bm{S_{\tau}} \cdot   \bm{T_{\tau}} \cdot \bm{S_{\hat{r}}}$ 
    & $\bm{T_{\hat{r}}}$\\
    \midrule
    V1 & $\bm{I}$ & $\bm{I}$ & $\bm{S_{\tau}} \cdot \bm{T_{\tau}} \cdot \bm{T_{\hat{r}}}$  \\
    
    V2
    & $\bm{I}$
    & $\bm{S_{\hat{r}}}$ 
    & $\bm{S_{\tau}} \cdot \bm{T_{\tau}} \cdot \bm{T_{\hat{r}}}$ \\
    V3 
    & $\bm{R_{\hat{r}}}$ 
    & $\bm{S_{\hat{r}}}$
    & $\bm{S_{\tau}} \cdot \bm{T_{\tau}} \cdot \bm{T_{\hat{r}}}$\\
    \midrule
    V4 
    & $\bm{I}$ 
    & $\bm{S_{\tau}} \cdot \bm{T_{\tau}} \cdot \bm{S_{\hat{r}}}$
    & $\bm{I}$ \\
    V5 
    & $\bm{R_{\hat{r}}}$ 
    & $\bm{S_{\tau}} \cdot \bm{T_{\tau}} \cdot \bm{S_{\hat{r}}}$
    & $\bm{T_{\hat{r}}}$\\
    \midrule
    V6
    & $\bm{S_{\tau}} \cdot \bm{T_{\tau}} \cdot \bm{R_{\hat{r}}}$
    & $\bm{I}$ 
    & $\bm{I}$\\
    V7
    & $\bm{S_{\tau}} \cdot \bm{T_{\tau}} \cdot \bm{R_{\hat{r}}}$
    & $\bm{S_{\hat{r}}}$
    & $\bm{T_{\hat{r}}}$\\
    \bottomrule
    \end{tabular}}
    \caption{Variant models of the first group.}
    \label{table:5}
\end{table} 
\begin{table}[ht]
\small
    \centering
    \renewcommand{\arraystretch}{1.5}
    \setlength{\tabcolsep}{0.5mm}{
    \begin{tabular}{c|c|c|c}
    \toprule
     & {$\bm{R_{\hat{r} \tau}}$ } &  {$\bm{S_{\hat{r} \tau}}$ } & {$\bm{T_{\hat{r} \tau}}$}  \\
    \midrule
    TCompoundE 
    & $\bm{I}$ 
    & $ \bm{S_{\tau}} \cdot   \bm{T_{\tau}} \cdot \bm{S_{\hat{r}}}$ 
    & $\bm{T_{\hat{r}}}$\\
    \midrule
    V8 
    & $\bm{I}$ 
    & $\bm{S_{\hat{r}}}$ 
    & $\bm{T_{\tau}} \cdot \bm{T_{\hat{r}}}$  \\
    V9
    & $\bm{I}$
    & $\bm{S_{\hat{r}}}$ 
    & $\bm{S_{\tau}} \cdot \bm{T_{\hat{r}}}$ \\
    V10 
    & $\bm{I}$ 
    & $\bm{S_{\hat{r}}}$
    & $\bm{R_{\tau}} \cdot \bm{T_{\hat{r}}}$\\
    V11 
    & $\bm{I}$ 
    & $\bm{S_{\hat{r}}}$
    & $\bm{\bm{R_{\tau}} \cdot \bm{S_{\tau}} \cdot \bm{T_{\tau}} \cdot \bm{T_{\hat{r}}}}$ \\
    \midrule
    V12 
    & $\bm{I}$ 
    & $\bm{T_{\tau}} \cdot \bm{S_{\hat{r}}}$
    & $\bm{T_{\hat{r}}}$\\
    V13
    & $\bm{I}$
    & $\bm{S_{\tau}} \cdot \bm{S_{\hat{r}}}$ 
    & $\bm{T_{\hat{r}}}$\\
    V14
    & $\bm{I}$
    & $\bm{R_{\tau}} \cdot \bm{S_{\hat{r}}}$
    & $\bm{T_{\hat{r}}}$\\
    V15
    & $\bm{I}$
    & $\bm{R_{\tau}} \cdot \bm{S_{\tau}} \cdot \bm{T_{\tau}} \cdot \bm{S_{\hat{r}}}$
    & $\bm{T_{\hat{r}}}$\\
    \bottomrule
    \end{tabular}}
    \caption{Variant models of the second group.}
    \label{table:6}
\end{table}

In keeping with section \ref{sec:diff}, the variant model is also divided into two groups, where the variant model in the first group keeps the time-specific operations unchanged, and changes the relation-specific operations; The second group keeps the relation-specific operations of the same and change the time-specific operations.
\begin{table*}[h]
\small
    \centering
    \begin{tabular}{c|ccc|cccc|cccc}
    \toprule
    \multirow{1}{*}{Mapping on} & \multicolumn{3}{c}{$\mathtt{Relation}$} & \multicolumn{4}{c}{$\mathtt{ICEWS14}$}& \multicolumn{4}{c}{$\mathtt{ICEWS05-15}$}  \\
    & $\bm{T_{\hat{r}}}$ & $\bm{S_{\hat{r}}}$ & $\bm{R_{\hat{r}}}$ & $\mathtt{MRR}$ & $\mathtt{H@1}$ & $\mathtt{H@3}$ & $\mathtt{H@10}$ & $\mathtt{MRR}$ & $\mathtt{H@1}$ & $\mathtt{H@3}$ & $\mathtt{H@10}$ \\
    \midrule
    $ \bm{S_{\hat{r}}}$  & \ding{52} & \ding{52} & & \textbf{0.644} & \textbf{0.561} & \textbf{0.694} & \textbf{0.795} &                \textbf{0.692} & \textbf{0.612} & \textbf{0.743} & \textbf{0.837}  \\
    \midrule
    \multirow{4}{*}{$\bm{T_{\hat{r}}}$}
    & \ding{52} & & & 0.304 & 0.128 & 0.412 & 0.632 & 
                                  0.374 & 0.194 & 0.490 & 0.702 \\
    & \ding{52} & \ding{52} & & 0.576 & 0.476 & 0.639 & 0.755                                & 0.623 & 0.521 & 0.690 & 0.809\\
    & \ding{52} & & \ding{52} & 0.547 & 0.442 & 0.612 & 0.740      & 0.606 & 0.522 & 0.655 & 0.760 \\
    & \ding{52} & \ding{52} & \ding{52} & 0.571 & 0.468 & 0.636 & 0.757          & 0.620 & 0.517 & 0.686 & 0.807 \\
    \midrule
    \multirow{3}{*}{ $\bm{S_{\hat{r}}}$}
    & & \ding{52} & & \underline{0.602} & \underline{0.518} & 0.650 & 0.762                            & \underline{0.676} & \underline{0.595} & \underline{0.727} & \underline{0.824} \\
    & & \ding{52} & \ding{52} & 0.580 & 0.478 & 0.644 & \underline{0.770}      & 0.649 & 0.556 & 0.713 & 0.818 \\
    & \ding{52} & \ding{52} & \ding{52} & 0.576 & 0.485 & 0.630 & 0.746          & 0.652 & 0.571 & 0.702 & 0.801 \\
    \midrule
    \multirow{4}{*}{ $\bm{R_{\hat{r}}}$}
    & & & \ding{52}  & 0.601 & 0.515 & \underline{0.651} & 0.761       
                                & 0.634 & 0.555 & 0.680 & 0.781 \\
    & & \ding{52} & \ding{52} & 0.587 & 0.491 & 0.644 & 0.763                                & 0.636 & 0.546 & 0.687 & 0.804\\ 
    & \ding{52} & & \ding{52} & 0.548 & 0.445 & 0.607 & 0.742                                & 0.606 & 0.522 & 0.655 & 0.760\\
    & \ding{52} & \ding{52} & \ding{52} & 0.582 & 0.491 & 0.638 & 0.752         & 0.638 & 0.550 & 0.690 & 0.804 \\
\bottomrule
    \end{tabular}
    \caption{Supplementary results of different relation-specific operations on ICEWS14 and ICEWS05-15}
    \label{table:l1}
\end{table*}
\begin{table*}[h]
\small
    \centering
    \begin{tabular}{c|ccc|cccc|cccc}
    \toprule
    \multirow{1}{*}{Mapping on} & \multicolumn{3}{c}{$\mathtt{Time}$} & \multicolumn{4}{c}{$\mathtt{ICEWS14}$}& \multicolumn{4}{c}{$\mathtt{ICEWS05-15}$}  \\
    & $\bm{T_{\tau}}$ & $\bm{S_{\tau}}$ & $\bm{R_{\tau}}$ & $\mathtt{MRR}$ & $\mathtt{H@1}$ & $\mathtt{H@3}$ & $\mathtt{H@10}$ & $\mathtt{MRR}$ & $\mathtt{H@1}$ & $\mathtt{H@3}$ & $\mathtt{H@10}$ \\
    \midrule
    
    $ \bm{S_{\hat{r}}} $ & \ding{52} & \ding{52} & & \textbf{0.644} & \textbf{0.561} & \textbf{0.694} & \textbf{0.795} &                \underline{0.692} & \textbf{0.612} & \underline{0.743} & \underline{0.837}  \\
    \midrule
    \multirow{6}{*}{$\bm{T_{\hat{r}}}$}
    & \ding{52} & & & 0.578 & 0.478 & 0.640 & 0.756 
                               & 0.621 & 0.519 & 0.685 & 0.807  \\
    & & \ding{52} & & 0.577 & 0.478 & 0.639 & 0.754        
                               & 0.620 & 0.517 & 0.685 & 0.807 \\
    & & & \ding{52} & 0.559 & 0.460 & 0.619 & 0.737        
                               & 0.584 & 0.482 & 0.646 & 0.771 \\
    & \ding{52} & & \ding{52} & 0.563 & 0.461 & 0.629 & 0.746                                & 0.601 & 0.494 & 0.670 & 0.795 \\
    & & \ding{52} & \ding{52} & 0.565 & 0.466 & 0.623 &   0.763                                & 0.607 & 0.505 & 0.672 & 0.793 \\
    & \ding{52} & \ding{52} & \ding{52} & 0.564 & 0.462 & 0.628 & 0.747                                  & 0.606 & 0.505 & 0.670 & 0.794 \\
    \midrule
    \multirow{6}{*}{$\bm{S_{\hat{r}}}$}
    & \ding{52} & & & 0.634 & 0.548 & \underline{0.687} & \underline{0.791}        
                               & \textbf{0.693} & \underline{0.607} & \textbf{0.750} & \textbf{0.846} \\
    & & \ding{52} & & \underline{0.638} & \underline{0.558} & 0.685 & 0.783         
                               & 0.682 & 0.606 & 0.735 & 0.832 \\
    & & & \ding{52} & 0.587 & 0.499 & 0.642 & 0.743         
                               & 0.639 & 0.557 & 0.688 & 0.786 \\
     & \ding{52} & & \ding{52} & 0.620 & 0.538 & 0.670 & 0.768                                & 0.680 & 0.598 & 0.734 & 0.826 \\
     & & \ding{52} & \ding{52} & 0.577 & 0.498 & 0.621 & 0.722                                & 0.671 & 0.592 & 0.720 & 0.814 \\
     & \ding{52} & \ding{52} & \ding{52} & 0.582 & 0.505 & 0.627 & 0.734                                  & 0.673 & 0.595 & 0.723 & 0.815  \\
\bottomrule
    \end{tabular}
    \caption{Supplementary results of different relation-specific operations on ICEWS14 and ICEWS05-15}
    \label{table:l2}
\end{table*}

From the first group of variant models introduced in Table \ref{table:5}, we can see that V1, V2 and V3 all apply time-specific operations to relation-specific translation operations; V4 and V5 apply time-specific operations to relation-specific scaling operations; V6 and V7 apply it to relationship-specific rotation operations.

A formula introduction for the second set of variant models can be viewed in Table \ref{table:6}. Where V8, V9, V10 and V11 all make time-specific operations specific to relation-specific translation operations; V12, V13, V14 and V15 apply it to relationship-specific scaling operations.

\section{Different Scoring Function} 
\label{apx:difscore}
The semantic similarity scoring function is shown in function \ref{fuc:sim}. The distance scoring function is defined as follows:
\begin{equation}\label{fuc:dis}
\phi (s,\hat{r},o, \tau) = \| \bm{e_{s}^{\hat{r} \tau}} - \bm{e_o} {\|}_2
\end{equation}

From the definition of the distance scoring function, we can see that in the distance scoring function, we use the L2 norm to find the Euclidean distance of $\bm{e_{s}^{\hat{r} \tau}}$ and $\bm{e_o}$.

\begin{table}[ht]
\small
    \centering
    \renewcommand{\arraystretch}{}
    \setlength{\tabcolsep}{2.5mm}{
    \begin{tabular}{c|c|c|c|c}
    \toprule
     & MRR &  H@1 & H@3 & H@10  \\
    \midrule
    Distance & 0.446 & 0.333 & 0.506 & 0.667 \\
    Similarity & \textbf{0.644} & \textbf{0.561} & \textbf{0.694} & \textbf{0.795}   \\
    \bottomrule
    \end{tabular}}
    \caption{Distance and similarity scoring function result on ICEWS14.}
    \label{table:score}
\end{table} 
We can observe that the semantic similarity scoring function is more advantageous in Table \ref{table:score}. In addition, the broadcast mechanism of the tensor can be better utilized by using the semantic similarity scoring function. It takes up less memory at runtime and has a shorter run time.

\section{More result of variant of TCompound}\label{Appedix:E}
In the experimental analysis section, we examined the utilization of single operations as well as variants incorporating all operations. This section presents additional experimental results, encompassing variants employing various combinations of two geometric operations, such as translation and rotation, and scaling and rotation. These combinations are applied to both relation-specific and time-specific operations. Our model consistently outperforms all aforementioned variants across all metrics on the ICEWS14 and ICEWS05-15 datasets, as illustrated in Tables \ref{table:l1} and \ref{table:l2}. These findings confirm the soundness of TCompoundE's design.

\section{Case Study} \label{apx:case}
We have selected some of the quadruples used in Section \ref{sec:sne} experiments as examples in the case study. The experimental results are shown in the table \ref{table:case}, where $\bm{e_s}$, $\bm{T_{\hat{r} \tau} e_s}$, and $\bm{S_{\hat{r} \tau} T_{\hat{r} \tau} e_s}$ respectively represent the three stages in Section \ref{sec:sne}.

\begin{table}[ht]
\small
    \centering
    \renewcommand{\arraystretch}{}
    \setlength{\tabcolsep}{5mm}{
    \begin{tabular}{c|c|c|c}
    \toprule
     & {$\bm{e_s}$ } &  {$\bm{T_{\hat{r} \tau} e_s}$ } & {$\bm{ S_{\hat{r} \tau}T_{\hat{r} \tau} e_s}$}  \\
    \midrule
    C1
    & 72
    & 38 
    & \textbf{1}\\
    C2
    & 6 
    & 4 
    & \textbf{1}  \\
    C3
    & 72
    & 38 
    & \textbf{1} \\
    C4
    & 7100
    & 7072
    & \textbf{2}\\
    C5
    & 431
    & 152
    & \textbf{3} \\
    C6
    & 87
    & 17
    & \textbf{1}\\
    C7
    & 65
    & 99 
    & \textbf{1} \\
    C8
    & 300
    & 9
    & \textbf{1}\\
    C9
    & 47
    & 41
    & \textbf{1}\\
    C10
    & 216
    & 105
    & \textbf{1}\\
    \bottomrule
    \end{tabular}}
    \caption{Case study on ICEWS14.}
    \label{table:case}
\end{table}

The table shows the ranking in each stage. We can observe from the table that the ranking of the sampled quadruples will have a greater improvement after the two stages than in the initial stage. The details of C1 - C10 are shown as follows: \\
\emph{(South Korea, Express intent to meet or negotiate, China, 2014-09-30) \\
(South Korea, Criticize or denounce, North Korea,2014-07-03)  \\
(South Korea, Express intent to meet or negotiate, China, 2014-09-30) \\
(South Korea, Host a visit, Barack Obama, 2014-04-23) \\
(South Korea, Express intent to cooperate, North Korea, 2014-03-27) \\
(South Korea, Express intent to cooperate, North Korea, 2014-03-27)\\
(South Korea, Host a visit, North Korea, 2014-10-08) \\
(South Korea, Express intent to cooperate, China, 2014-07-20)\\
(South Korea, Make an appeal or request, China, 2014-06-25)\\
(South Korea, Consult, Head of Government (Egypt), 2014-11-23)\\
}


\end{document}